\documentclass{article}

\usepackage{template}

\usepackage[utf8]{inputenc} 
\usepackage[T1]{fontenc}    
\usepackage{hyperref}       
\usepackage{url}            
\usepackage{booktabs}       
\usepackage{multirow}
\usepackage{array}
\usepackage{makecell}
\usepackage{amsmath,amssymb}
\usepackage{bm}
\usepackage{amsfonts}       
\usepackage{nicefrac}       
\usepackage{microtype}      
\usepackage{cleveref}       
\usepackage{lipsum}         
\usepackage{graphicx}
\usepackage{natbib}
\usepackage{doi}
\usepackage{bbding}

\newtheorem{theorem}{Theorem}[section]

\newtheorem{corollary}[theorem]{Corollary}
\newtheorem{definition}[theorem]{Definition}

\newtheorem{remark}[theorem]{Remark}

\title{Rethinking the Zigzag Flattening for Image Reading}

\author{ 
    {\hspace{1mm}Qingsong Zhao}\\
	Tongji University, China\\
	\texttt{qingsongzhao@tongji.edu.cn} \\
	\And
	{\hspace{1mm}Zhipeng Zhou} \\
	Chinese Academay of Sciences, Beijing, China \\
	\texttt{zhouzhipeng113@mails.ucas.ac.cn  } \\
	\And
    {\hspace{1mm}Yi Wang} \\
	Shanghai AI Laboratory, China\\
	\texttt{wangyi@pjlab.org.cn} \\
	\And
    {\hspace{1mm} Yu Qiao} \\
	Shanghai AI Laboratory, China\\
	\texttt{qiaoyu@pjlab.org.cn} \\
 	\And
    {\hspace{1mm} Limin Wang} \\
	Nanjing University, China\\
	\texttt{lmwang.nju@gmail.com} \\
	\And
      {\hspace{1mm} Duoqian Miao } \\
	Tongji University, China\\
	\texttt{dqmiao@tongji.edu.cn } \\
        \And
    {\hspace{1mm} Cairong Zhao \Envelope} \\
	Tongji University, China\\
	\texttt{zhaocairong@tongji.edu.cn } \\
}

\renewcommand{\shorttitle}{Hilbert Flattening: a Locality-Preserving Matrix Unfolding Method }

\hypersetup{
pdftitle={},
pdfsubject={q-bio.NC, q-bio.QM},
pdfauthor={Tsingsong Zhao},
pdfkeywords={First keyword, Second keyword, More},
}
\usepackage{xcolor}

\begin{document}
\maketitle

\begin{abstract}
Sequence ordering of word vector matters a lot to text reading, which has been proven in natural language processing (NLP). 
However, the rule of different sequence ordering in computer vision (CV) was not well explored,
e.g., why the ``zigzag" flattening (ZF) is commonly utilized as a default option to get the image patches ordering in vision networks.
Notably, when decomposing multi-scale images, the ZF could not maintain the invariance of feature point positions. 
To this end, we investigate the Hilbert fractal flattening (HF) as another method for sequence ordering in CV and contrast it against ZF. 
The HF has proven to be superior to other curves in maintaining spatial  locality, when performing multi-scale transformations of dimensional space.
And it can be easily plugged into most deep neural networks (DNNs).
Extensive experiments demonstrate that it can yield consistent and significant performance boosts for a variety of architectures.
Finally, we hope that our studies spark further research about the flattening strategy of image reading.
\end{abstract}

\section{Introduction}
Humans usually read text by row or by column, but how do you ``read" a 2D image?
We first look at the area of greatest interest and then the other areas or patches.
And how the DNNs do with the image and text?
Arguably, regardless of the text or image, many DNNs read it as text. 
Recently years, inspired by the Transformer extension successes in NLP, Convolution-free architectures, in particular MLP-Mixer \cite{MLP-Mixer}, have become the model of choice in computer vision.

To apply Transformer from NLP to CV, however, the image embedding scale had to be reduced on account of computational cost of the Transformers would scale quadratically with the number of pixels.
An image can simply be treated as $16\times16$ words in MLP-Mixer, but is it suitable to utilize a text-reading approach to read the images?
In this work, we investigated the question of whether, fundamentally, the ``zigzag'' flattening strategy (aka line by line) is the optimal solution for image reading. 
Intuitively, the writing can be compared to a one-way time line, which ranked the key information. 
On the contrary, the vision system projects the static real world onto a two-dimensional screen, which constitutes the image. 

\begin{figure}[!t]
\begin{center}
\includegraphics[width=120mm]{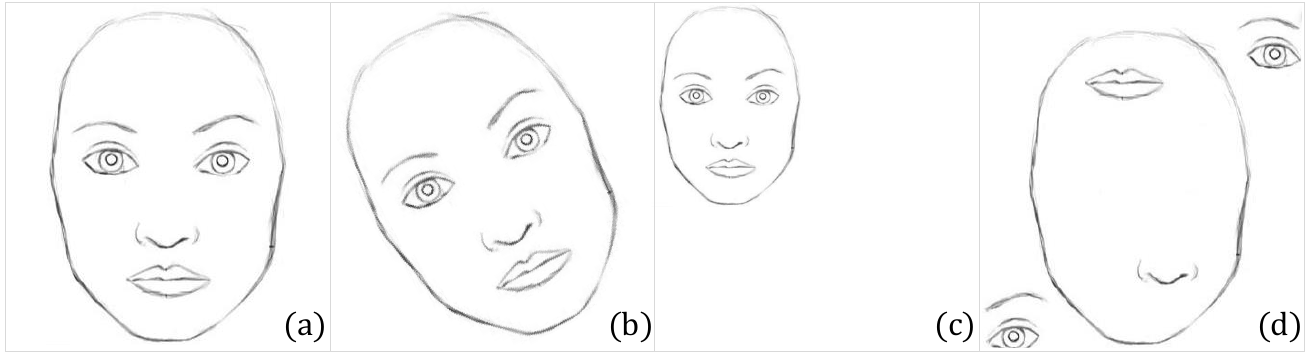}
\end{center}
  \caption{Illustration of Image Reading.
  }
\label{fig:face}
\end{figure}

The key information on the image can be considered as an undirected graph \cite{bronstein2021geometric}. 
In other words, all the key information on an image can not be represented in a one-way vector.
This also explains that the semantics expressed by the image is unchanged after rotation, see Fig. \ref{fig:face} (a) and (b). In addition, the position of the key information in the text must be fixed.
If it was moved, the original message conveyed by the text would be subverted.
This also explained that why the position encoding is prominent for Transformer to capture sequence ordering of input tokens.
However, the key information of the image which consists of whole blocks of pixels would not change in any way even if it was moved by panning or scaling, see Fig. \ref{fig:face} (a) and (c).
Notably, the feature points inside an image block can not be moved, just like the position of a keyword in a sentence.
If the features were moved, the semantics expressed by the image would change radically, see Fig. \ref{fig:face} (a) and (d).

In the nutshell, there is a fundamental difference in the modality text and images represent information.
Hence, it is naive to apply the mode of reading text directly to read images in CV. 
For this purpose, we investigated the Hilbert fractal flattening strategy (aka Pseudo-Hilbert curve flattening or ``Hilbert” for short) as another method for image reading in CV and contrast it against ZF. 
The Hilbert is the only space-filling curve whose Hausdorff–Besicovitch dimension is greater than its topological dimension \cite{albers2008benoit}.
And it has been shown to outperform the other curves in remaining the spatial locality, when transforming from a multi-dimensional space to a one-dimensional space \cite{tkde_analysis_hilbert}.  
Several related works have applied it simply to the indexing of image pixels in CV.
But, its theoretical explanation and application potential have not been well investigated, and even remains controversial.

In this paper, we first discussed the nature of the Hilbert fractal in image dimensional transformation and its scale robustness.
Then, extensive experiments including Dynamic Time Warping (DTW) distance, interpolation based image resize, image classification, etc, demonstrated that Hilbert flattening
was a better image reading method compared to ``zigzag" flattening.
Our contributions can be summarized as follows: 
\begin{itemize}
\item 
We posed a simple but easily taken-for-granted question.
In MLP-Mixer, does the model have to use the same paradigm for reading images as it does for text reading?
Through theoretical analysis and fine experimental design, we have attempted to give one answer and to generalize this question to a larger context.
\item 
We answered previous controversial questions.
We have theoretically estimated the square-to-linear dilation factor of the finite approximation of Hilbert curve.
This indicates that the consecutive parts in sequence are close in the corresponding image, and explains  that why \cite{hilbert_curve_for_sEMG} can feed the 1D signals to the CNNs.
Meanwhile, The Average Square Distance was proposed to give a quantitative description of comparison between inverse Hilbert flattening and inverse Zigzag flattening on probability of points close in 2-dimension are close in linear sequence.
In addition, we empirically demonstrate that the Hilbert flattening can maintain feature consistency in multi-scale images.
\item 
We proposed a new patch embedding method, named Hilbert Patch Embedding (HPE), dedicated to any DNNs, considering both effectiveness and simplicity.
The HPE are simple and can be easily plugged into most DNNs. Experiments demonstrate that, without introducing additional hyperparameters, it can improve MLP-Mixer and the proposed Feature Pyramid Network (FPN-MLPs) by $1.2\%$ (Top1 Acc) and $ 4.29\%$ over their original models on CIFAR-10 \cite{krizhevsky2014cifar}, respectively.
\end{itemize}

\section{Related Works}
In this section, we first overview the applications of Hilbert curves respectively according to their motivations.
Then, we provide a review of the development of MLP-Only architectures,
which serve as the backbones of our experimental section.

\subsection{Applications of Hilbert Curves}
Such prominent works in the field of Mathematics as \cite{linear_clustering,tip96space_filling_curves,tkde_analysis_hilbert} have evidenced that the locality between objects in multi-dimensional space is preserved in linear space. 
Inspired by such idea, recent widely works consist of \cite{hilbert_curve_for_sEMG, tip19hybrid_LSTM, thinking_in_patch} have been proposed to introduce the Hilbert curves into a CV application.
\cite{tip19hybrid_LSTM} noted that the order of the image patches has a significant impact on the performance of the Long-Short Term Memory (LSTM), 
and if the zigzag flattening was performed in the horizontal direction, the neighboring blocks in the vertical direction are far apart.
Eventually, the LSTM may not establish the connection between those patches well.
To improve the performance of localization in the detection of image forgeries, 
they utilized the Hilbert curves to arrange image patches before the block sequences were fed into the LSTM.
With the same idea, to extract the better spatial features, FDPT \cite{thinking_in_patch} also utilized the Hilbert curves to flatten image patches before feed them into the Gated Recurrent Unit (GRU).
By contrast, \cite{hilbert_curve_for_sEMG} employed the Hilbert curves to generate 2D image representations from 1D surface electromyography (sEMG) signals, then the features of the sEMG signals were extracted by the CNN based backbones. 
But, the above methods only apply the Hilbert curve to a CV task without in-depth theoretical analysis and fine empirical experimental proofs.

\subsection{MLPs}
\cite{MLP-Mixer} proposed a new architecture named MLP-Mixer that differs from CNNs and Transformers by eliminating the need for convolution and self-attention, which relies only on the repeated implementations of MLPs across the spatial or feature channels.
Those works all employed Zigzag flattening to expand 2-D images or features into 1-D patch or token sequences.
With the same motivation, ResMLP \cite{ResMLP} exploited the effect of data augmentation and knowledge distillation on training a MLPs based architecture.
Those works above all employed the zigzag flattening to expand 2D images or features into 1D patch or pixel sequences.
But the ZF would move the initially adjacent image blocks (semantically related patches) away from each other, but HF does not, see Fig. \ref{fig:zvshilbert} for details.
Hence, in this paper, we explored Hilbert curves whose cluster property outperform zigzag curves for those MLPs based architectures.

\begin{figure*}[!tb]
\begin{center}
\includegraphics[width=1\linewidth]{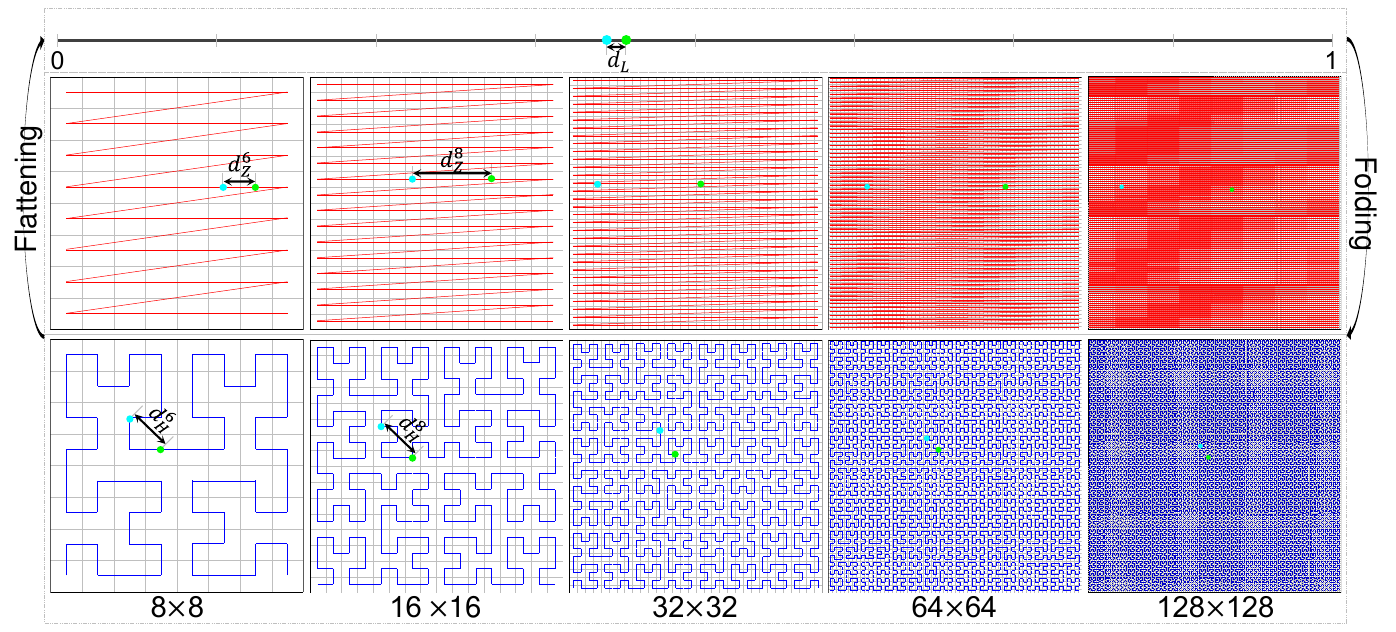}
\end{center}
  \caption{Multi-scale transformation of dimensional space with Zigzag curve and Hilbert curve flattening/folding, respectively. 
  Specifically, suppose that two points separated by $d_L \in (0^+, 1)$ are taken at random on the real number axis in the interval $I = [0,1]$, 
  which can always be taken if $2^n\times 2^n $ ($ n \in \{3,4,5,6,7\}$) points are equally spaced on the number axis.
  By dividing the interval of the numerical axis by $2^6$ equal parts and using different unfolding operations (ZF and HF), 
  we find that the distance $d_L$ between two points changes to $d_Z^6$ and $d_H^6$.
  Similarly, if $2^8$ equal divisions are performed, the distance between the two points again changes to $d_Z^8$ and $d_H^8$.
  The two points with fixed distance $d_L$ on $[0,1]$ mapped to 2-D space under different types of unfolding operations will have various distances.
  }
\label{Hilbert-and-Zigzag-curve}
\end{figure*}

\section{Hilbert Flattening}  \label{flattening}
The proposed Hilbert flattening is inspired by the Hilbert curve of space-filling curves (SFC) from the geometric theory of \textit{Fractals} \citep{sprecher2002space}, 
just like Zigzag one mimics how people read text from left to right, row by row. 
Hilbert curves preserve the most locality of the input features among all candidates when flattening such a multi-dimensional matrix input into a 1-D vector.
In this section, we first introduce the definition of SFC and how it can be introduced into image unfolding operations.
Next, we analyze the properties of HF, in the comparison with existing unfolding operations.
With the property analysis of HF, we show how to apply it to popular models in practical applications.

\subsection{Preliminaries}
\paragraph{Space-Filling Curves} \label{SFCs}
A continuous curve is called \textit{space-filling curve} if it can pass through every point of a closed square \cite{peano1890courbe}.
More precisely, a space-filling curve is a continuous mapping from a closed unit interval $I=[0,1]$ to a closed unit square $Q=[0,1]^2$ \cite{simmons1963introduction}.
It is defined as follows:
\begin{definition}
A mapping $f$: $I\rightarrow E^n(n\geq 2)$ is continuous and $f(I)$ has positive Peano–Jordan measure, then $f(I)$ is called a space-filling curve, where $E^n$ denotes an n-dimensional Euclidean space.
\end{definition}
Hilbert curve \cite{hilbert1935stetige} is the first generic geometric program that allows to construct entire classes of space-filling curves.
Compared to zigzag curves and Gray-encoded curves, Hilbert curves were the best at minimizing the number of clusters \cite{linear_clustering}.
More precisely, as shown in Fig. \ref{Hilbert-and-Zigzag-curve}, a space-filling curve  \citep{simmons1963introduction} is a continuous mapping from a closed unit interval $I=[0,1]$ to a closed unit square $[0,1]\times[0,1]$.
The definitions and known theorems used in this paper mainly come from \cite{sagan2012space}. 

\paragraph{Hilbert Curve}
We assume $\mathcal{I}$ and $\mathcal{Q}$ as the interval $[0,1]$ and square $[0,1]\times [0,1]$ respectively.
The generating process of the Hilbert curve is driven by the following: 
\begin{equation}
\begin{aligned}
\mathcal{H}:t& \in[0,1]\mapsto \mathcal{H}(t)\in[0,1]\times [0,1],\\
t&=0.q_{1}q_{2}\cdots, 0\le q_{j}\le 3,\\
\mathcal{H}(t)&=\left(
    \begin{aligned}
    &\mathcal{R}e\\
    &\mathcal{I}m,
    \end{aligned}\right)\lim\limits_{n\rightarrow \infty}T_{q_{1}}T_{q_{2}}\cdots T_{q_{n}}\mathcal{Q},
\end{aligned}
\end{equation}
where $t$ is represented in quaternary form. 
The definition of $\{T_{i}|0\le i\le 3\}$ is defined as follows:
\begin{equation}
    \begin{aligned}
    &T_{i}z=\frac{1}{2}H_{i}z+h_{i}, 0\le i\le 3,\\
    &\begin{aligned}
    &H_{0}z=\bar{z}i,H_{1}z=z,H_{2}z=z,H_{3}z=-\bar{z}i,\\
    &h_{0}=0,h_{1}=\frac{i}{2},h_{2}=\frac{1+i}{2},h_{3}=\frac{2+i}{2},
    \end{aligned}
    \end{aligned}
\end{equation}
where we consider complex numbers $z\in\mathbb{C}$ as $(Re(z),$ $ Im(z))\in \mathcal{Q}$.
The transformations $\{T_{i}|0\le i\le 3\}$ defined above correspond to different geometric deformations.
Take transformation $T_{0}$ as an example, we first shrink the original $\mathcal{Q}$ towards the original point under the ratio $\frac{1}{2}$, 
then reflect on the imaginary axis by multiplying with $-1$ and rotate the square through $90^{\circ}$ by multiplying with imaginary number $i$.

During the generating process of the Hilbert curve, the sub-squares shrink into points, which claims that $\mathcal{H}(t)$ is a point in $\mathbb{R}^{2}$.
We construct the $n$-th approximation of the Hilbert curve by
$n$-th iteration, which is denoted as $\mathcal{H}_{n}$, 

\begin{equation}\label{Hilbert Formula}
\begin{aligned}
    \mathcal{H}_{n}(0.q_{1}q_{2} \cdots q_{n})&=\left(
    \begin{aligned}
    &\mathcal{R}e\\
    &\mathcal{I}m
    \end{aligned}
    \right)\sum\limits_{j=1}^{n}\frac{1}{2^{j}}H_{q_{0}}H_{q_{1}}H_{q_{2}}\cdots H_{q_{j-1}}h_{q_{j}},\\
    &=\sum\limits_{j=1}^{n}\frac{1}{2^{j}}(-1)^{e_{0j}}\text{sgn}(q_{j})\left(
    \begin{aligned}
    &(1-d_{j})q_{j}-1\\
    &1-d_{j}q_{j}
    \end{aligned}
    \right)\\
    \text{sgn}(x)&=\left\{
    \begin{aligned}
    &1, \text{ if } x>0,\\
    &0, x=0.
    \end{aligned}
    \right.\\
    e_{kj}&=\#(\text{ ''k" preceding } q_{j})\mod 2,\\
    d_{j}&=e_{0j} + e_{3j} \mod 2,
\end{aligned}
\end{equation}

where $\#$ is the counting function and $k\in \{0,3\}$.
We have drawn the image points of finite quaternary ($2\le n\le 3$) connected by straight lines in Figure flattening curves left.
Note that the order $n$ approximation of the Hilbert curve originates in the lower-left sub-square
and terminates in the lower-right sub-square.
The exit point from each sub-square coincides with the point that goes into the following sub-square. 

\paragraph{Zigzag curve}
We study the Zigzag curve on the image with a size of $H\times W$.
For convenience, we assume that both $H$ and $W$ are equal to 1 and divided uniformly into $2^{n}$ parts.
Given a real number $t\in [0,1]$ which can be represented in quaternary form with finite length:
$t=0.q_{1}q_{2}\cdots q_{n}$,
it is defined by $\mathcal{Z}$ as follows:
\begin{equation}
    \mathcal{Z}:0.q_{1}q_{2}\cdots q_{n} \mapsto 
    \left(
    \begin{aligned}
    & (\sum\limits_{k=1}^{n}q_{k}4^{n-k}\% 2^{n}) * \frac{1}{2^{n}} + \frac{1}{2^{n+1}}\\
    & \lfloor\frac{\sum\limits_{k=1}^{n}q_{k}4^{n-k}}{2^{n}}\rfloor* \frac{1}{2^{n}}+\frac{1}{2^{n+1}}
    \end{aligned}
    \right)
\end{equation}

\paragraph{Morton Curve}
The generating process of the Morton curve is similar to the Hilbert curve's. 
We first denote the $n$-th approximation of the Morton curve as $\mathcal{M}_{n}$.
Morton curve is the limit of $\mathcal{M}_{n}$ as $n$ goes to infinity.
The conversion function from binary to decimal is denoted by $\mathcal{B}$.
The generating process of the Morton curve is driven as follows:
\begin{equation}\label{morton_curve}
\begin{aligned}
    &\mathcal{M}_{n}: t\in\mathcal{I}\mapsto \mathcal{M}_{n}(t)\in \mathcal{Q}\\
    &\mathcal{M}_{n}(\frac{\mathcal{B}(q_{1}q_{2}\cdots q_{n})}{2^{n}-1})=\left(
    \begin{aligned}
        & \frac{\mathcal{B}(q_{1}q_{3}\cdots q_{n})}{2^{n/2}}\\
        & \frac{\mathcal{B}(q_{2}q_{4}\cdots q_{n-1})}{2^{n/2}}
    \end{aligned}
    \right)\triangleq p\in\mathcal{Q}\\
    & \mathcal{M}_{n}(t)= p_{0}* (1-s)+p_{1}*s, t\in [\mathcal{M}_{n}^{-1}(p_{0}), \mathcal{M}_{n}^{-1}(p_{1})]\\
    & \text{where }s=\frac{t-\mathcal{M}_{n}^{-1}(p_{0})}{\mathcal{M}_{n}^{-1}(p_{1})-\mathcal{M}_{n}^{-1}(p_{0})}\in[0,1], q_{i}\in\{0,1\}\\
\end{aligned}
\end{equation}

\begin{figure}[!t]
\begin{center}
\includegraphics[width=0.80\textwidth]{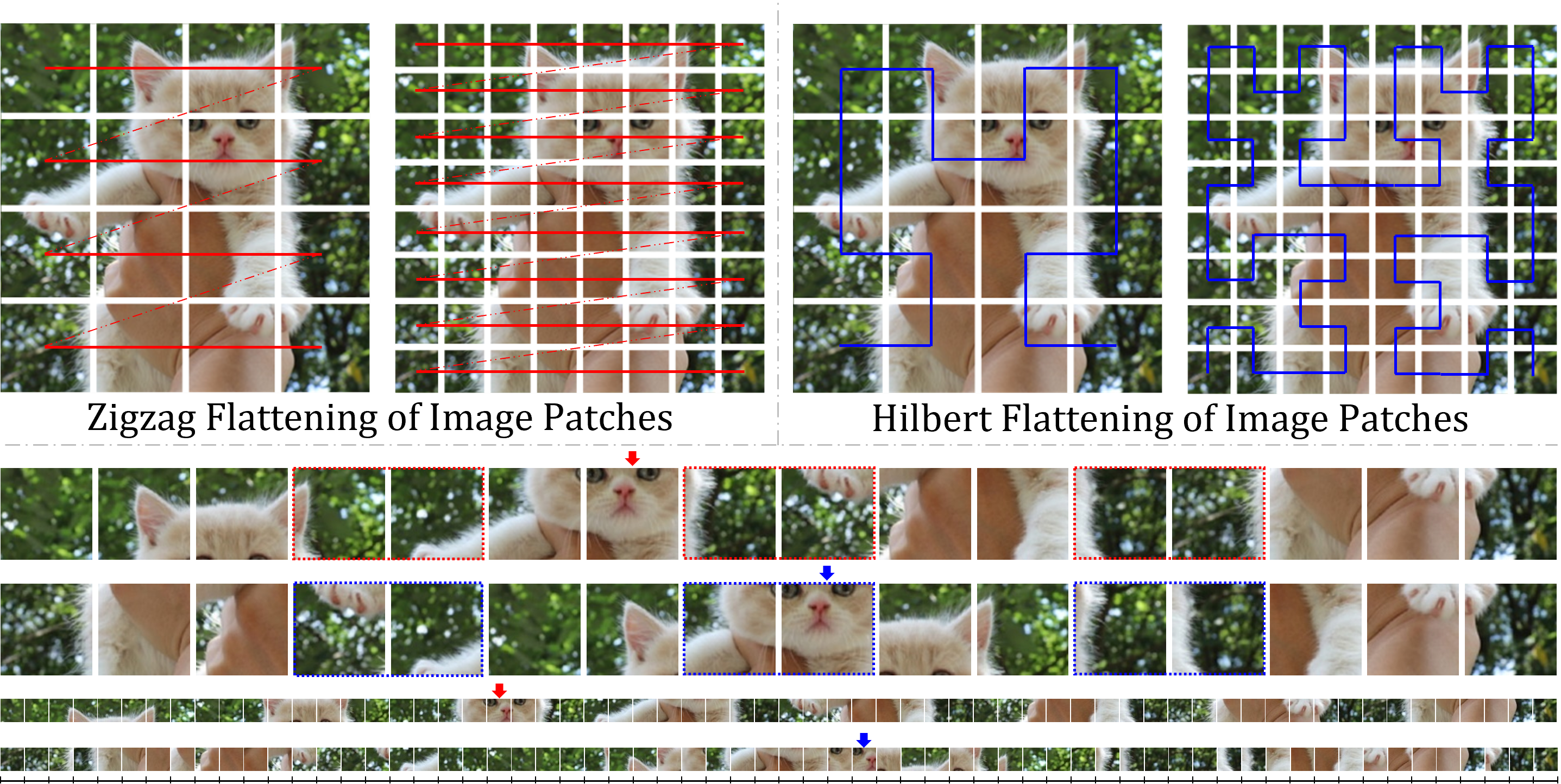}
\end{center}
  \caption{Zigzag flattening (red lines and arrows) VS. Hilbert flattening (blue lines and arrows) in ViT for image patch expanding. 
  When flattening a 2D image into a 1D patch sequence, ZF will move the initially adjacent image blocks (semantically related patches) away from each other, but HF does not.
  That is, the head of cat remained clustered together after slicing at different flattening scales, and the position of the head on the 1D sequence was not change.
}
\label{fig:zvshilbert}
\end{figure}

\paragraph{Matrix Unfolding} \label{MF}
In computer vision, the matrix unfolding operation takes place when a tensor undergoes a dimensional change.
For example, as illustrated in Fig. \ref{fig:zvshilbert}, the flattening of 2-dimensional grid data into 1-dimensional linear data is a matrix unfolding operation.
We can observe that it is precisely the inverse of SFC operation.
In this paper, three SFCs will be introduced, namely Hilbert curves, Morton curves, and Zigzag curves.
Each of the three SFCs corresponds to a specific matrix flattening method: HF for Hilbert curves, MF for Morton curves, and ZF for Zigzag curves.
The Hilbert curves \citep{hilbert1935stetige} is the first generic geometric program that allows the construction of entire classes of space-filling curves.
The Morton curves (aka, Z-order \citep{leccons}) map multidimensional data to one dimension while preserving the locality of the data points in data structures. 
And, the Zigzag curve is the default operation for flattening tensors in deep learning (i.e., \textit{torch.flatten()}).


We give the expressions for the three matrix unfolding methods as follows.

\paragraph{Hilbert Flattening}
Hilbert flattening is built upon the Hilbert curve,
with the approximation of the Hilbert curve of order $n$,
we defined its operation. 
Consider an image with resolution $n\times n$,
the inverse map of the approximation of Hilbert curve at order $n$ provides the mechanism of HF:
\begin{equation}
\mathcal{H}_{n}^{-1}:
    \left(
    \begin{aligned}
    & \frac{i}{2^{n}}+\frac{1}{2^{n+1}}\\
    &\frac{j}{2^{n}}+\frac{1}{2^{n+1}}
    \end{aligned}
    \right)\mapsto =0.q_{1}q_{2}\cdots q_{n},
\end{equation}
where $\mathcal{H}_{n}(0.q_{1}q_{2}\cdots q_{n})=(\frac{i}{2^{n}}+\frac{1}{2^{n+1}},\frac{j}{2^{n}}+\frac{1}{2^{n+1}})^{T}$.
Then the pixel on the image contains point $(\frac{i}{2^{n}}+\frac{1}{2^{n+1}},\frac{j}{2^{n}}+\frac{1}{2^{n+1}})^{T}$ will be assigned the value $0.q_{1}q_{2}\cdots q_{n}$.
All the pixels on the images will be ordered by their values, which in fact gives the definition of Hilbert Flattening.

\paragraph{Zigzag Flattening}
Same as HF, given a real number $t\in [0,1]$ which can be represented in quaternary form with finite length:
$t=0.q_{1}q_{2}\cdots q_{n}$,
the ZF is defined by $\mathcal{Z}^{-1}$ as follows:
\begin{equation}
\mathcal{Z}^{-1}: [\frac{i}{2^{n}},\frac{j}{2^{n}}]\mapsto 0.q_{1}q_{2}\cdots q_{n}=\mathcal{Z}^{-1}([\frac{i}{2^{n}},\frac{j}{2^{n}}]),
\end{equation}
where 
$0\le i,j\le 2^{n}-1$.

\paragraph{Morton Flattening}
According to the definition of Morton Curve in Equation (\ref{morton_curve}), we define the Morton Flattening of order $n$ by 
\begin{equation}
\mathcal{M}_{n}^{-1}:\left(
\begin{aligned}
    & \frac{\mathcal{B}(q_{1}^{0}q_{2}^{0}\cdots q_{n-1}^{0})}{2^{n/2}} \\
    & \frac{\mathcal{B}(q_{1}^{1}q_{2}^{1}\cdots q_{n-1}^{1}}{2^{n/2}} \\
\end{aligned}
\right)\mapsto \frac{\mathcal{B}(q_{1}^{0}q_{1}^{1}\cdots q_{n-1}^{0}q_{n-1}^{1})}{2^{n}-1}
\end{equation}
where $\mathcal{M}_{n}^{-1}$ is the inverse mapping of $\mathcal{M}_{n}$.

\subsection{Properties} \label{Properties} 
We analyzed locality preserving and scale robustness about the three aforementioned unfolding methods using toy examples, respectively.
Specifically, by preserving the 2-D geometric structure in a 1-D format, we believe the HF can enhance MLP-like architectures in visual discriminations, as evidenced in Section \ref{classification}.

\subsubsection{Locality Preserving}  \label{long_range_preservation}
Theoretically, we demonstrate that HF maximizes the preservation of 2-D topological structure among the other flattening methods,
notably surpassing the orthodox Zigzag method.
This holds true whether flattening 2-D matrixes into 1-D vectors or folding 1-D vectors into 2-D matrixes, confirming symmetry in dimension reduction or expansion. 

\paragraph{Flattening 2-D to 1-D} 
Flattening techniques can transform high-dimensional data such as matrices or tensors into flattened representations that can be processed by neural networks, highlighting its significance in deep learning. 
Indeed, the transformation process inevitably disrupts the original grid structure.
For instance, when applying the ZF operation, only the neighboring pixel points within each row maintain their adjacency,
while the structural coherence across rows is lost. 
To facilitate the learning of locality representations in images,
it is essential to ensure the local smoothness of the input data matrix.
This concept often referred to as locality bias, is a widely adopted assumption in computer vision. 
And the locality bias assumes that nearby pixels or regions in an image exhibit similar characteristics or share common patterns. 
By retaining the spatial relationships between pixels or regions, 
the model can better capture local dependencies and patterns, 
leading to improved performance in various computer vision tasks \citep{VVT, nguyen2020wide}.

\begin{figure}[!tb]
\begin{center}
\includegraphics[width=1.0\linewidth]{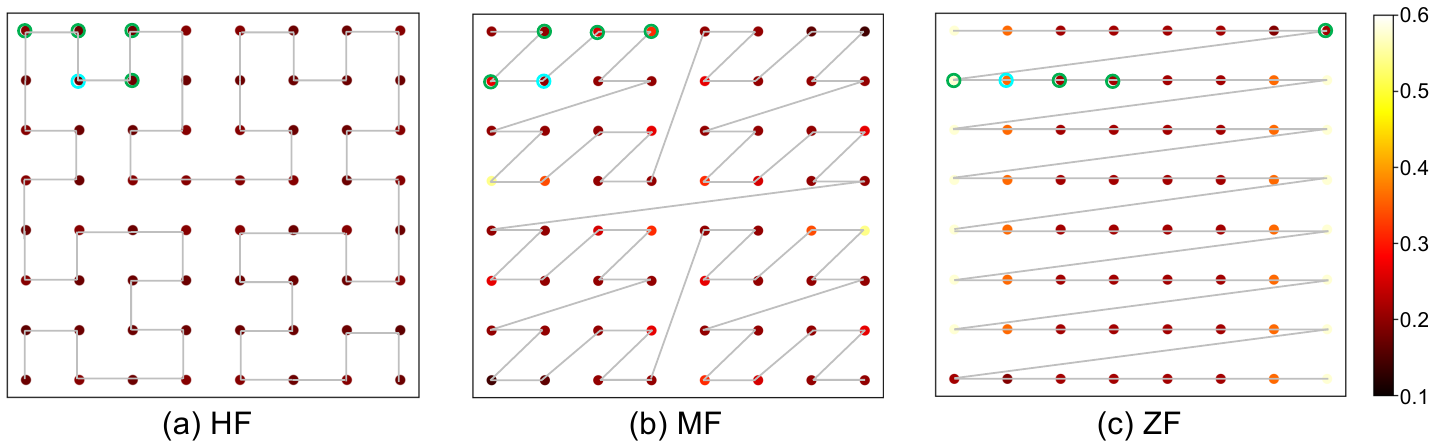}
\end{center}
  \caption{
  Suppose we compute the DeGrid at each point on a $8 \times8$ grid structure data with $K=2$.
  The results are expressed as a heat map, in which the brightness of the pixels indicates the grid structure info preservation. 
  The closer the pixels are to black, the better.}
\label{DeGrid}
\end{figure}

\begin{table}[!tb]
\caption{
The percentage of grid structure preservation with the given DeGrid threshold $\epsilon$, where we set $K=2$. 
Indexing pixel points within the same 2-D neighborhood, the larger the percentage indicates the better structure preservation. 
}
\begin{center}
\scalebox{0.66}{
\begin{tabular}{|l|c|c|c|c|c|c|c|c|c|c|c|c|c|c|}
\toprule[1pt]
$\epsilon$  &0.043&0.045 &0.047  &0.053 &0.055 &0.057 &0.059 &0.069 &0.070 &0.071 &0.077 &0.086 &0.087 &0.120\\   
\toprule[0.5pt]
HF   &60.16\%  &100.00\% &100.00\%   &100.00\% &100.00\% &100.00\% &100.00\% &100.00\% &100.00\% &100.00\% &100.00\% &100.00\% &100.00\% &100.00\%\\ 
MF     &0.39\%  &0.39\% &62.89\%   &62.89\% &62.89\% &69.14\% &69.14\% &81.64\% &81.64\% &87.89\% &91.02\% &91.02\% &92.58\% &92.58\%\\
ZF     &0.00\%  &0.20\% &0.20\%   &87.89\% &87.89\% &87.89\% &87.89\% &87.89\% &87.89\% &87.89\% &87.89\% &87.89\% &87.89\% &87.89\%\\ 
\toprule[1pt]
\end{tabular}}
\end{center}
\label{DeGrid_table}
\end{table}

Now we explore how much the flattened 1-D sequence can preserve the original 2-D structure info.
As depicted in Fig. \ref{DeGrid}, the process begins by selecting a neighborhood with a radius of $K=2$ from the flattened 1-D data.
Within this neighborhood, a central node $p$ is identified, i.e., the points circled in cyan.
The degree of deformation of this neighborhood is determined by calculating the sum of the squares of the distances between each pixel (i.e., the points circled in green) and the central node $p$ over the 2-D grid data.
This measure provides insight into how much the neighborhood has been altered or transformed.
To describe the degree of destruction of the grid structure within this neighborhood,
the ratio of the degree of deformation to the sum of the elements in the neighborhood is computed, we call it the Deformation Value of 2-D Grid Structure.

\paragraph{Formal Definition of DeGrid}
The steps for calculating the DeGrid have been given in Section \ref{Properties} \textit{Grid Structure Preservation},
and we give its formal expression as follows.
Suppose we have a sequence of points of length $N$.
For each point $p$ at position $i$ $(1\le i \le N)$,
we collect the neighbors which are $K$ steps away from $p$.
The DeGrid definition of these pixel points with respect to $p$ is
\begin{equation}
\textbf{DeGrid}(p)=\frac{\sum\limits_{\genfrac{}{}{0pt}{2}{i-K \le k \le i+K}{1\le p+k\le N}}\|\mathcal{F}^{-1}(p_{k})-\mathcal{F}^{-1}(p)\|_{2}}{\#(K \text{ step neighbors})}.
\end{equation}
A higher DeGrid indicates a lower degree of retention of the grid structure, while a lower ratio suggests a less significant deformation or disruption of the original grid structure.

\begin{table}[!tb]
\caption{Theoretical derivation of the dilation factor and limits for the three space-filling curves.
}
\begin{center}
\resizebox{0.6\linewidth}{!}{
\begin{tabular}{|l|l|l|}
\toprule[1pt]
Flattening Methods & Dilation Factor Lower Bound & Limits \\ 
\toprule[0.5pt]
Zigzag curves & $4^n-2^{n+1}+2$ & $+\infty$ \\ 
\toprule[0.5pt]
Morton curves & $2^{n}-2^{-n}$ & $+\infty$ \\ 
\toprule[0.5pt]
Hilbert curves & $6$ (refer to \cite{bauman2006dilation})& $ 6 $ \\  
\toprule[1pt]
\end{tabular}}
\end{center}
\label{dalition_factors}
\end{table}

Lastly, as presented in Table \ref{DeGrid_table}, 
we establish a threshold for the DeGrid. 
Nodes with a DeGrid below this threshold $\epsilon$ are deemed to have a manageable deformation. 
The grid structure preservation within the DeGrid threshold is determined by the ratio between the total number of these points and the total number of nodes in the flattened 2-D grid data.
Indeed, as illustrated in Fig. \ref{DeGrid}, it is evident that when the $\epsilon$ exceeds $0.045$, the grid structure preservation of the HF can surpass that of other SFCs entirely.

\paragraph{Folding 1-D to 2-D} 
When performing a folding operation on a 1-D vector,
it leads to an expansion of the spatial distance between the original data points,
i.e., loss of local-range correlation info.
As depicted in Fig. \ref{Hilbert-and-Zigzag-curve},
when transforming 1-D sequence data into a 2-D grid using unfolding methods,
the distance between two points (i.e., $d_L$) is expanded,
resulting in an increased spatial distance of $d_Z$ and $d_H$, respectively.
We want the process to increase $d_L$ by as small a factor (i.e., $d_{\cal F}/{d_L}$, where ${\cal F}$ denotes a matrix flattening method) as possible to preserve the locality info.
Similar to \cite{estevez2015visualizing}, 
we formulate this dilation factor (DF) to describe the preservation of the 1-D local-range correlation information.

For elaboration, let's consider folding the interval $I=[0,1]$ into a $2^n \times 2^n$ grid data. 
As depicted in Fig. \ref{Hilbert-and-Zigzag-curve}, 
when using the ZF method,
the original distance between two points is indefinitely and continuously enlarged as the dimension of the 2-D grid increases.
Instead, that distance converges to a definite value with the HF method. 
In Table \ref{dalition_factors}, 
we also provide the dilation factor expressions for the above three flattening methods,
along with their corresponding limit values.
We find that as $n$ tends to positive infinity, 
only the limit of HF is a constant (i.e., the $\mathbb{C}$ is equal to 6 \citep{bauman2006dilation}) for all three expansion methods.

The properties of different matrix flattening have been discussed in above,
where we derived some qualitative remarks.
Here, we give formal definitions and proofs of these claims as follows.

\paragraph{Dilation Factor and Limits} 
We introduce a definition of the limit, namely the dilation factor.
Specifically, given two points $t_{1},t_{2}\in[0,1]$, 
the quanternary form are represented as $t^{1}=0.q_{1}^{1}q_{2}^{1}\cdots$ and $t^{2}=0.q_{1}^{2}q_{2}^{2}\cdots$ when these two points are close. It means that for an large integer $j$ such that $q_{k}^{1}=q_{k}^{2},\forall 1\le k\le j$.
By applying the formula in Equation \ref{Hilbert Formula}, we obtain the distance between points of $\mathcal{H}(t^{1}),\mathcal{H}(t^{2})$ as follows:
\begin{equation}
    |\mathcal{H}(t^{1})-\mathcal{H}(t^{2})|^{2}\le \sum\limits_{k=j+1}\frac{8}{2^{k}}\le \frac{8}{2^{j}}.
\end{equation}
The dilation bound of the Hilbert curve is shown in Theorem \ref{Dilation_HF},
We find that HF operation can obtain a sequence ordering of the image/feature map which guarantees that consecutive parts in sequence are close in the original image.

Now we study the dilation factor of the ZF operation on an image with a size of $H\times W$.
Let 
$t^{1}=0.\underbrace{00\cdots 0}_{\frac{n}{2}}\underbrace{33\cdots 3}_{\frac{n}{2}}$ and $t^{2}=0.\underbrace{00\cdots 0}_{\frac{n}{2}-1}1\underbrace{00\cdots 0}_{\frac{n}{2}}$, which are consecutive points in the interval $[0,1]$ with distance $\frac{1}{4^{n}}$.
We have $\frac{|\mathcal{Z}(t^{1})-\mathcal{Z}(t^{2})|^{2}}{\frac{1}{4^{n}}}=\frac{(1-\frac{1}{2^{n}})^{2}+\frac{1}{4^{n}}}{\frac{1}{4^{n}}}=4^{n}-2^{n+1}+2$.
Then we get Remark \ref{ZF_dia}.
And, our proposed Remark has been cited in the point cloud classification and segmentation tasks \citep{chen2022efficient}.

{
\small 
For Morton flattening, let
$p_{0}=\left(
\begin{aligned}
& \mathcal{B}(0\underbrace{1\cdots1}_{\frac{n}{2}-1})/2^{n/2}\\
& \mathcal{B}(\underbrace{1\cdots1}_{\frac{n}{2}})/2^{n/2}\\
\end{aligned}\right)$
,$p_{1}=\left(
\begin{aligned}
& \mathcal{B}(1\underbrace{0\cdots0}_{\frac{n}{2}-1})/2^{n/2}\\
& \mathcal{B}(\underbrace{0\cdots0}_{\frac{n}{2}})/2^{n/2}\\
\end{aligned}\right)$,
}
we have $\mathcal{M}_{n}^{-1}(p_{0})=\frac{2^{n/2}}{2^{n}-1}$ and $\mathcal{M}_{n}^{-1}(p_{1})=\frac{2^{n/2}-1}{2^{n}-1}$. 
So we have 
\begin{equation}
    \frac{|p_{0}-p_{1}|^{2}}{|\mathcal{M}_{n}^{-1}(p_{0})-\mathcal{M}_{n}^{-1}(p_{1})|}=\frac{\frac{1}{2^n}+1}{\frac{1}{2^{n}-1}}=2^{n}-2^{-n}.
\end{equation}
Then we get Remark \ref{MF_dia}.
\begin{theorem}\label{Dilation_HF}
The square-to-linear \textbf{dilation factor} of the Peano-Hilbert curve is equal to 6 \citep{article}, which means that the maximum value of $\frac{|\mathcal{H}(t^1)-\mathcal{H}(t^2)|^{2}}{|t^1-t^2|}\le 6$. 
\end{theorem}
\begin{remark}\label{ZF_dia}
The square-to-linear dilation factor of the ZF curve is $\infty$. ($\lim\limits_{n\rightarrow\infty}4^n-2^{n+1}+2=\infty$).
\end{remark}
\begin{remark}\label{MF_dia}
The square-to-linear dilation factor of the MF curve is $\infty$. ($\lim\limits_{n\rightarrow\infty}2^{n}-2^{-n}=\infty$).
\end{remark}

As mentioned in Section \ref{Properties} \textit{Scale Robustness of Folding}, 
as $n \rightarrow \infty$, 
we employ the value of the ratio between the distances after multiscale folding to indicate the scale robustness. 
Below we give the steps for calculating the ratios for HF and ZF, respectively.

\begin{equation}
    \begin{aligned}
        \lim\limits_{n\rightarrow \infty}\frac{d_{H}^{2n}}{d_{H}^{2(n+1)}}=\lim\limits_{n\rightarrow \infty}\frac{d_{H}^{2n}}{d_{L}}\lim\limits_{n\rightarrow \infty}\frac{d_{L}}{d_{H}^{2(n+1)}}=6\times \frac{1}{6}=1
    \end{aligned}
\end{equation}

\begin{equation}
    \begin{aligned}
        \lim\limits_{n\rightarrow \infty}\frac{d_{Z}^{2n}}{d_{Z}^{2(n+1)}}=\lim\limits_{n\rightarrow \infty}\frac{d_{Z}^{2n}}{d_{L}}\lim\limits_{n\rightarrow \infty}\frac{d_{L}}{d_{Z}^{2(n+1)}}=\lim\limits_{n\rightarrow \infty}\frac{4^{n}-2^{n+1}+2}{4^{n+1}-2^{n+2}+2}=\frac{1}{4}
    \end{aligned}
\end{equation}

\begin{figure}[!tb]
\begin{center}
\includegraphics[width=0.65\linewidth]{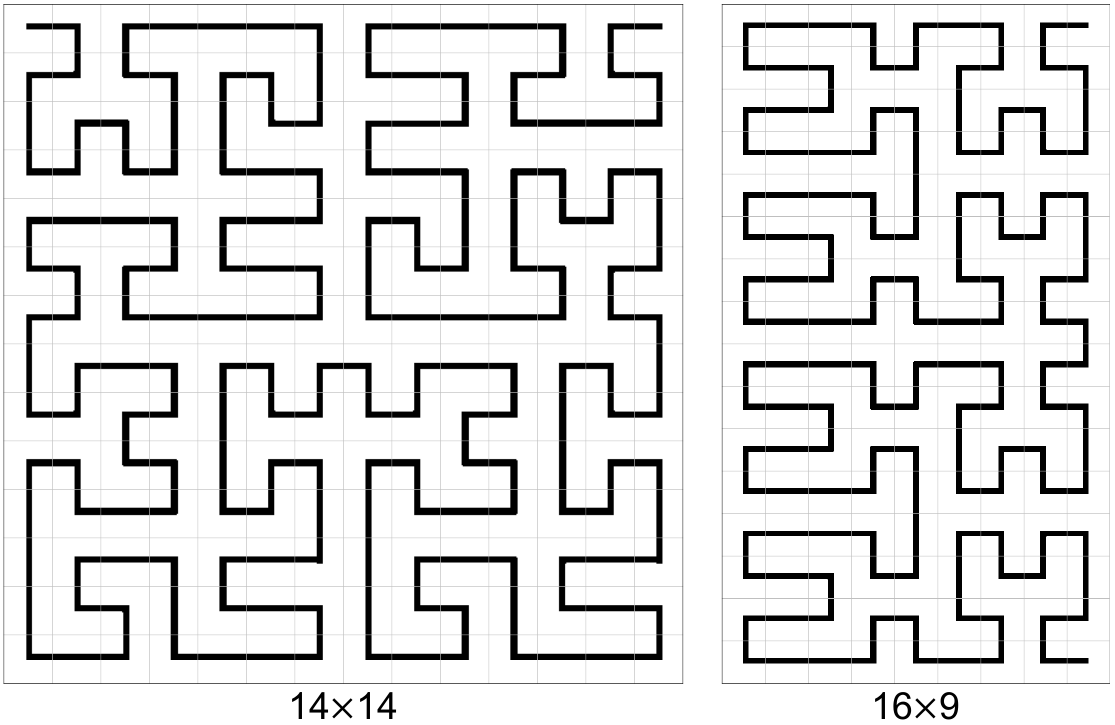}
\end{center}
\caption{
    Thanks to \citep{pseudo, arbitrary}, we present here examples of arbitrary Hilbert flattening. The left panel shows a resolution of $14\times14$, while the right panel at $16\times9$.
}
\label{fig:hilbert_arb}
\end{figure}

\paragraph{Symmetry Group Definition of Scale Robustness}
According to the general equivariant \citep{bronstein2021geometric,wang2022filter} of the convolution operator which is defined in Definition \ref{G_robust},
we give a corollary about $\mathcal{S}$-robust of the flattening operator.
Take $\Omega=\mathbb{Z}_{2^n}\times \mathbb{Z}_{2^n}$ to be a 2-D grid, and $\Omega'=\mathbb{Z}_{2^{n+n}}$ to be a 1-D sequence.
Consider the $n$-th order and $(n+1)$-th order approximation of Hilbert flattening as reported in Eq. \ref{Hilbert Formula}, geometrically, 
the HF operation just divides the $n$-th order approximation Hilbert curve uniformly between every pair of endpoints into three parts, 
then moves the second part away from the original curve with distance $\frac{1}{2^{n+1}}$.
Finally, it connects the moving part with the endpoints of the second part
(please turn to Fig. \ref{Hilbert-and-Zigzag-curve} for details).

Given an image $I$ with size $2^{n+1}\times 2^{n+1}$,
we utilize the $(n+1)$-th order HF to unfold it.
We denote the pixel set after flattening as $\mathcal{H}_{n+1}(I)$.
Also, 
we first scale down the image $I$ into image $I_{{1}/{2}}$ with size $2^{n}\times 2^{n}$.
We denote the pixel set after $n$-th order HF as $\mathcal{H}_{n}(I_{{1}/{2}})$.
According to the previous two paragraphs,
$\mathcal{H}_{n}(I_{{1}/{2}})$ and $\mathcal{H}_{n+1}(I)$
satisfy the following condition:
\begin{equation} \label{robust_eq12}
    (\mathcal{H}_{n+1}(I))_{{1}/{2}}\approx \mathcal{H}_{n}(I_{{1}/{2}}),
\end{equation}
where ${{1}/{2}}$ means that image scaling ratio. 
Consider the scale operation group $\mathcal{S}=\{(\cdot)_{2^{-m}}|m\in \mathbb{Z}\}$,
we have 
\begin{equation}
(\mathcal{F}_{n+m}(I))_{2^{-m}}\approx \mathcal{F}_{n}(I_{2^{-m}}),
\end{equation}
where $\mathcal{F}$ is a flattening operator, and we get the Corollary \ref{S_robust}. 
In conclusion, as $n$ approaches a sufficiently large number (i.e., $n \rightarrow +\infty$), we \textbf{only} find the Hilbert flattening is $\mathcal{S}$-robust.
\begin{definition} \label{G_robust}
A function $f:\mathcal{X}(\Omega)\rightarrow \mathcal{X}(\Omega)$ is $\mathcal{G}$-robust if $f(\rho(g)x)\approx\rho(g)f(x)$ for all $g\in \mathcal{G}$, i.e., group action on the input affects the output in the same way, where $\rho $ is a representation of group $\mathcal{S}$,  $\mathcal{X}(\Omega)$ denotes all signals on domain $\Omega$.
\end{definition}

\begin{corollary} \label{S_robust} 
A flattening function $\mathcal{F}:\mathcal{X}(\Omega)\rightarrow \mathcal{X}(\Omega')$ is $\mathcal{S}$-robust if $\mathcal{F}(\rho(g)x) \approx \rho'(g)\mathcal{F}(x)$ for all $g\in \mathcal{S}$, i.e., group action on the input affects the output in the same way with input and output spaces having different domains $\Omega, \Omega'$  and representations $\rho, \rho'$ of the same group $\mathcal{S}$.
\end{corollary}


\subsection{Arbitrary Hilbert Flattening}
For clarity, we focus on the grid with equal size of height and width ($2^n $).
As shown in Fig. \ref{fig:hilbert_arb}, 
following a simple recursive algorithm proposed in \citep{arbitrary, pseudo},
the general Hilbert flattening can cover the grid with arbitrary size.

\section{Experiments}
In this section, two analysis experiments were first reported to compare the scale robustness of ZF and HF. 
Second, a FPN-MLPs architecture was proposed to compare the multi-scale representations stability of both. 
Third, We introduce an implementation of patch embedding for the MLP-only architectures, see Fig. \ref{fig:zvshilbert}, which sets the flattening strategy between image patches. 

\paragraph{Experimental Setup}
We utilized the common settings to compare the performance for fairness.
With limited computational resources, 
we are not motivated by practice-based CV tasks.
Notably, 
the settings including software (i.e., the virtual environment of Python 3.7 and Torch 1.7.1) and hardware (i.e., 4$\times$ NVIDIA 1080Ti GPUs) are strictly consistent.

\subsection{Image Scaling.}
\label{subsubsec:Image Scaling}
\begin{figure}[!tb]
\begin{center}
\includegraphics[width=120mm]{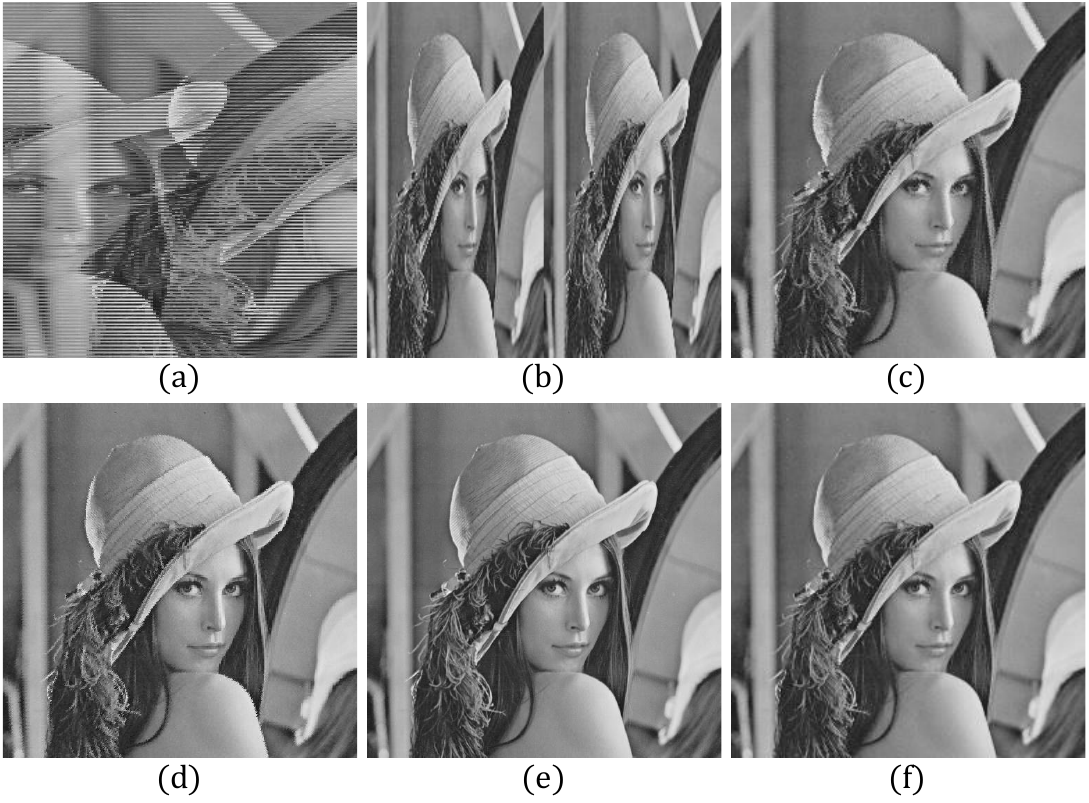}
\end{center}
  \caption{Qualitative results of the 1D interpolation-based image scaling. 
  (a) After expanding the image from 2D to 1D with ZF, we downsample the original image to $256\times{256}$ by the nearest neighbor 1D interpolation algorithm. 
  (b) Again, we do the dimensional transform with ZF first, upsamle the original image to $512\times512$ by the same algorithm. 
  (c) Same as (a), but with HF. 
  (d) Same as (b), but with HF. 
  (e) $512\times512$ resolution original ``Lena" image. 
  (f) $256\times256$ resolution original image.
  }
\label{fig:resize}
\end{figure}

Image scaling is a common operation in digital image processing (DIP). 
Two interpolations by row and column respectively are the standard practice for image scaling. 
But, if we expand the image and interpolate it only once, will this scale the image properly? 
As shown in Fig. \ref{fig:resize} (a) and (b), with ZF method, neither up-sampling nor down-sampling operations result in a normal image. 
On the contrary, with HF strategy, see Fig. \ref{fig:resize} (c) and (d), the normal results are obtained regardless of the upsampling or downsampling operation. 
Moreover, the scaling effect of HF can perform favorably against the result of 2D interpolation algorithms.

\begin{figure}[!tb]
\begin{center}
\includegraphics[width=0.8\linewidth]{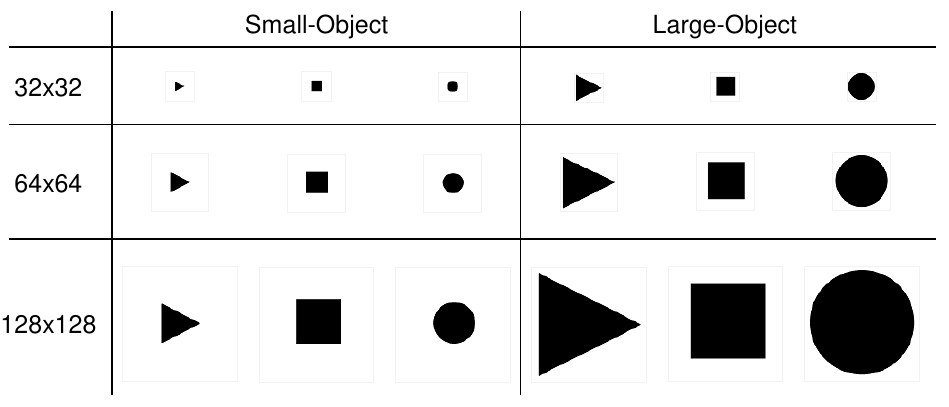}
\end{center}
  \caption{ A toy dataset owning 18 synthetic images to measure scale robustness empirically.
  }
\label{toy_images}
\end{figure}

\subsection{Scale Robustness of Folding}
The scale robustness of folding refers to the extent to which the relationship between two chosen points changes in different folding scales. Generally, we suppose HF is robust to the folding scale while ZF is \textbf{not}.
Referring to Fig. \ref{Hilbert-and-Zigzag-curve}, assume that the interval $I$ is folded into $2^n \times 2^n$ grid data again.
As $n$ increases, if the distance $d_{\cal F}^{2n}$ between the two selected points converges to a nonzero constant value (i.e., $ \lim\limits_{n\rightarrow\infty}{d_{\cal F}^{2n}} \in \mathbb{C}^{+}$),
which makes the ratio (i.e.,  $ \lim\limits_{n\rightarrow\infty}{d_{\cal F}^{2n}}/{d_{\cal F}^{2(n+1)}}$) between the distances after multiscale folding is $1$,
we suppose this folding process is robust to scale.
Specifically, when performing a multiscale folding operation using the ZF, the ratio between the two distances becomes indeterminate, i.e., $ \lim\limits_{n\rightarrow\infty}{d_Z^{2n}}/{d_Z^{2(n+1)}} \neq 1, s.t., \lim\limits_{n\rightarrow\infty}{d_Z^{2n}} = \infty $.
In contrast, when applying the same operation using the HF, that ratio tends to become convergent, i.e., $ \lim\limits_{n\rightarrow\infty}{d_H^{2n}}/{d_H^{2(n+1)}} = 1, s.t., \lim\limits_{n\rightarrow\infty}{d_H^{2n}} = 6$.

\begin{table*}[!tb]
\caption{
The DTW distance for different flattening methods on the proposed toy dataset (consisting of multiple scale object and image resolutions), Lower is better.
As presented in Fig. \ref{toy_images}, ``L32" means Large scale object with a resolution of $32\times32$ and ``S128" means Small scale one with a resolution of $128\times128$.
}
\begin{center}
\scalebox{0.75}{
\begin{tabular}{|l|c|c|c|c|c|c|c|c|c|c|c|c|c|c|c|c|c|c|}
\toprule[1pt]
Scale & \multicolumn{3}{c|}{ L32 vs S32 }   &  \multicolumn{3}{c|}{ L32 vs L64 }  &  \multicolumn{3}{c|}{ L64 vs S64 }  &  \multicolumn{3}{c|}{ L64 vs L128 }  & \multicolumn{3}{c|}{ L128 vs S128 } & \multicolumn{3}{c|}{ L32 vs S128 } \\ 
\cline{2-19} 
Methods   & HF & MF & ZF   & HF & MF & ZF  & HF & MF & ZF  & HF & MF & ZF  & HF & MF  & ZF & HF  & MF  & ZF \\ 
\toprule[0.5pt] 
Circle    & \textbf{5.19} & 5.71 & 8.61  & \textbf{4.13} &4.97 & 14.32  & \textbf{6.14} & 7.86 & 16.28  & \textbf{6.74} & 7.29 &28.71   & \textbf{9.13} & 10.20 & 33.31 & \textbf{3.95} & 4.38 & 18.52 \\ 
Square    & \textbf{3.40} & 4.06 & 8.14  & \textbf{5.42} &5.49 & 17.16  & \textbf{6.60} &7.26 & 15.53  & \textbf{7.39} &7.61 & 35.47  & \textbf{10.58} &12.23 & 31.03 & 8.67 &\textbf{7.61} & 29.67 \\ 
Triangle  & 4.61 &\textbf{4.59} & 9.60  & \textbf{3.63} &4.27 & 16.03  & \textbf{6.93} &6.99 & 18.83  & \textbf{6.06} &6.64 & 31.55  & \textbf{7.99} &9.37  & 38.07 & \textbf{4.30} & 4.76 & 10.28  \\
\toprule[1pt]
\end{tabular}}
\end{center}
\label{table:dtw}
\end{table*}

\begin{table}[!bt]
\caption{On CIFAR-10, recognition accuracy of different flattening methods with the proposed FPN-MLPs. ``Residual-H-Backbone-Z" indicates that the flattening approaches in the residual branch and backbone are HF and ZF, respectively. Same for the other settings.}
\begin{center}
\scalebox{1.2}{
\begin{tabular}{lll}
\toprule[1pt]
Flattening Methods   &Top-1\%  &Top-5\% \\ 
\toprule[0.5pt]
Residual-Z-Backbone-Z  & 81.42 & 99.18 \\
Residual-H-Backbone-Z  & 85.45 & 99.54 \\
Residual-Z-Backbone-H  & 81.67 & 99.14 \\
Residual-H-Backbone-H  & \textbf{85.71} & 99.58 \\
\toprule[1pt]
\end{tabular}}
\end{center}
\label{table:fpn_mlp_cifar10}
\end{table}

\label{dtw}

We show how ZF, MF, and HF respond to flattening 2-D toy examples in scale robustness. 
A toy dataset comprising three common shapes (\textit{circles, squares, and triangles}) with two scales (small and large) is constructed, as illustrated in Fig. \ref{toy_images}. 
Further, 
to compute the correlation between different 1-D folding vectors,
dynamic time warping (DTW) distance \citep{berndt1994using, dtw2020tcsvt} is employed as it allows for the comparison and alignment of sequences with variable lengths. 
The computed DTW distances for various sequence pairs are presented in Table \ref{table:dtw}, 
providing insights into the pairwise dissimilarity of the flattened images.
When the resolution of image pairs is increased, the DTW distance between large and small targets also increases. 
This observation indicates that higher resolution amplifies the differences between multi-scale targets within an image.
It is worth noting that the variation in HF is relatively small when comparing the DTW distance of different strategies.
This suggests that the HF is more robust to resolution changes and preserves the structural consistency of the flattened sequences.

\subsection{Image Classification} 
\label{classification}

\begin{figure}[!tb]
\begin{center}
\includegraphics[width=0.85\linewidth]{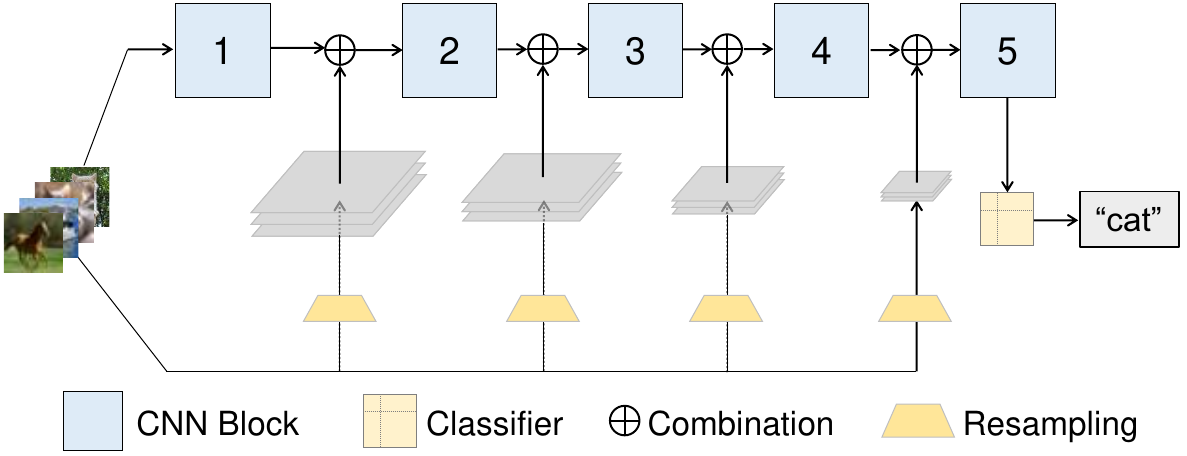}
\end{center}
  \caption{ 
  The illustration of the proposed MLP-FPN.
  } 
\label{fpn_mlp}
\end{figure}

We apply HF to the MLP-only models, constructing advanced MLP-Mixer and MLP-FPN.
As depicted in Fig. \ref{fpn_mlp}, the HF can be readily applied in various deep-learning operations whenever there is a need for matrix resampling or alignment,
e.g., it was employed in the patch embedding module.

\begin{table}[!bt]
 \caption{The outline of the proposed network architecture FPN-MLPs. 
  The output size of each block is the input size of the next one see Fig. \ref{fpn_mlp}. 
  From top to bottom, the components appear in sequence. 
  Each component may appear multiple times in FPN-MLPs.
}
\begin{center}
\scalebox{1.2}{
\begin{tabular}{ccc}
\toprule[1pt]
\makecell{Type} & \makecell{Patch size/Stride\\ or Remarks} & \makecell{Input Size} \\ 
\toprule[0.5pt]
Conv1D      & $7\times1/1$   & $3\times1024$ \\
Conv1D      & $5\times1/2$   & $64\times256$ \\
Conv1D      & $3\times1/2$   & $256\times64$ \\
Conv1D      & $3\times1/2$   & $512\times16$ \\
Conv1D      & $3\times1/1$   & $3\times1024/256/64/16$ \\
Conv1D      & $1\times1/1$   & $3\times1024/256/64/16$ \\
GELU        & $3\times1/1$   & $3\times1024/256/64/16$ \\
LayerNorm   & $3\times1/1$   & $3\times1024/256/64/16$ \\
AvgPool1D   & $16\times1 $   & $512\times16$ \\
Linear      & $Logits    $   & $1\times512$  \\
Softmax     & $Classifier$   & $1\times10$  \\ 
\toprule[1pt]
\end{tabular}}
\end{center}
\label{table:fpn_mlp}
\end{table}

\paragraph{MLP-FPN}
Intuitively, we believe that the scale robustness property of HF can enhance the multi-scale representation of the network containing a feature pyramid network (FPN) branch, 
resulting in improved performance gains.
To this end, we devised a network architecture solely based on the MLP module, while incorporating an FPN branching structure, namely MLP-FPN as presented in Fig. \ref{fpn_mlp}.
The FPN was originally introduced to address the challenge of multi-scale variation in object detection \citep{lin2017feature,min2022attentional}.
It accomplishes this by effectively integrating features that possess strong semantic information at lower resolutions and features with abundant spatial details at higher resolutions.
Inspired by this,
The residual branch of our MLP-FPN is formed by stacking down-sampling pyramid modules,
which include our HTS algorithm and 1-D convolutional neural networks (Conv1-D). 
The MLP consisting of Conv1-D is then assembled into the backbone network.
The outline of the proposed network architecture FPN-MLPs is shown in Table \ref{table:fpn_mlp}.

These experiments involved the validation of the proposed FPN-MLP network, 
supporting the hypothesis that HTS can enhance the multi-scale representation of the networks. 
The results are presented in Table \ref{table:fpn_mlp_cifar10}, we see that the MLP-Mixer with the HTS achieves clear gaps over the baseline (e.g., 1.2\% on MLP-Mixer-B/4).
We also find the FPN-MLP using the HTS can outperform the baseline by an obvious margin (i.e., 4.29\% earnings).

\paragraph{MLP-Mixer}
Both convolution-free and attention-free model MLP-Mixer, which is an architecture based exclusively on MLPs. 
Similar to ViT \cite{ViT}, it also treats an image as $16 \times 16$ words. 
That is, the patch embedding was an essential component of this model. 
As shown in Table \ref{table:mixer_cifar10}, we conduct numerous experiments with different patch embedding approaches on CIFAR-10 by utilizing the MLP-Mixer. 
The results note that the proposed HF is effective in MLP-Mixer and also achieves significant improvement based on the original patch embedding method. 
In addition, a 1D convolution based patch embedding method was proposed to compare HF and ZF. 
The experiments demonstrated that there is a obvious gap between the best accuracy of HF and ZF.
Notably, when we utilized the overlap convolution, this gap was widened.

\begin{table}[!t]
  \caption{Recognition accuracy of different patch embedding methods on CIFAR-10.
  ``Inter-Z Intra-H" means that the flattening strategy in the inter-patches and intra-patches are ZF and HF, respectively. 
  Same for the other settings.
  ``Conv1D-H" indicates that encoding the image patches by the 1D convolution, and the flattening method of image patches is HF.
  ``Conv1D-Z" means that the flattening method of image patches is ZF.
  ``Overlap" indicates that the kernel size of Conv1D is bigger than the patch size.
}
\begin{center}
\scalebox{1.2}{
\begin{tabular}{lcccc}
\toprule[1pt]
Methods & Patch Size & Overlap  &Top-1\%  &Top-5\% \\ 
\toprule[0.5pt]
Original    & 8 & -  & 87.05 & 99.15 \\
Inter-H Intra-Z         & 8 & -  & \textbf{87.35} & 99.19 \\
Inter-Z Intra-Z         & 8 & -  & 86.74 & 99.16 \\
Inter-Z Intra-H        & 8 & -  & 86.86 & 99.08 \\
Inter-H Intra-H         & 8 & -  &86.75 & 99.18 \\
\toprule[0.5pt]
Conv1D-Z    & 8 & -  &83.58 & 98.71 \\
Conv1D-H    & 8 & -  &\textbf{84.52} & 98.83 \\
\toprule[0.5pt]
Conv1D-Z    & 4 & -  &79.73 & 98.34 \\
Conv1D-H    & 4 & -  &\textbf{80.59} & 98.55 \\
Conv1D-Z  & 4 & $\checkmark$  &80.48 & 98.32 \\
Conv1D-H  & 4 & $\checkmark$  &\textbf{81.68} & 98.57 \\
\toprule[1pt]
\end{tabular}}
\end{center}
\label{table:mixer_cifar10}
\end{table}

\section{Conclusion}
In this paper, we explored whether Hilbert flattening is a better fit for image reading than Zigzag flattening. 
To answer the above question, we theoretically evaluate the square-to-linear dilation factor of the finite approximation of Hilbert curve, and propose the Average Square Distance to compare inverse HF with ZF. 
Based on the above theory, we proposed a new patch embedding method for MLPs.
Extensive experiments including dynamic time warping distances, interpolation based image resize, and image classification demonstrate that HF is more effective than ZF.
The code will be released soon.


\bibliographystyle{unsrtnat}
\bibliography{template}


\end{document}